\documentclass[journal]{IEEEtran}
\usepackage{cite}
\usepackage{times}
\usepackage{epsfig}
\usepackage{graphicx}
\usepackage{amsmath}
\usepackage{amssymb}

\usepackage{epsfig}
\usepackage{graphicx}
\usepackage{amsmath}
\usepackage{amssymb}
\usepackage{array}
\usepackage{tablefootnote}
\usepackage{subfigure}
\usepackage{float}
\usepackage{caption}
\usepackage{setspace}
\usepackage{hyperref}
\hypersetup{hypertex=true,
	colorlinks=true,
	linkcolor=blue,
	anchorcolor=blue,
	citecolor=blue}
\usepackage{algorithm}
\usepackage{algpseudocode}
\usepackage{graphics}

\usepackage{pifont}

\usepackage{multirow}

\usepackage{booktabs}
\usepackage{threeparttable}

\usepackage{amsfonts,amssymb} 

\usepackage[square,sort&compress,numbers]{natbib}

\usepackage{amsmath}

\usepackage{epstopdf}

\usepackage{makecell}

\usepackage{pifont}

\usepackage{soul}

\usepackage{graphicx}
\usepackage{amsmath}
\usepackage{amsfonts,amssymb}
\usepackage{booktabs}
\usepackage{threeparttable}
\usepackage{makecell}
\usepackage{multirow}
\usepackage{pifont}

\usepackage{xcolor}
\usepackage{soul}
\soulregister\cite7
\soulregister\eqref7
\allowdisplaybreaks
\begin{document}

%%%%%%%%% TITLE
\title{Complementary Frequency-Varying Awareness Network for Open-Set Fine-Grained Image Recognition}

%% omit the following lines up until the closing ``}''.
%% Additional authors and addresses can be added with ``\and'',
%% just like the second author.
%% To save space, use either the email address or home page, not both
%\and

%	Institution1\\
%	Institution1 address\\
%	{\tt\small firstauthor@i1.org}
%	\and
%	Second Author\\
%	Institution2\\
%	First line of institution2 address\\
%	{\tt\small secondauthor@i2.org}
%}

\author{Qiulei Dong, Jiayin Sun and Mengyu Gao% <-this % stops a space
	\thanks{The corresponding author is Qiulei Dong.
		
%		Jiayin Sun and Qiulei Dong are with the National Laboratory of Pattern Recognition, Institute of Automation, Chinese Academy of Sciences, Beijing 100190, China, the School of Artificial Intelligence, University of Chinese Academy of Sciences, Beijing 100049, China, and the Center for Excellence in Brain Science and Intelligence Technology, Chinese Academy of Sciences, Beijing 100190, China (e-mail: jiayin.sun@nlpr.ia.ac.cn; qldong@nlpr.ia.ac.cn).
%		
%		Hong Wang is with the College of Life Science, University of Chinese Academy of Sciences, Beijing 100049, China (email: hwang@ucas.ac.cn)
	}
}

%\markboth{Submitted to IEEE TRANSACTIONS ON CIRCUITS AND SYSTEMS FOR VIDEO TECHNOLOGY}%

\maketitle
%\ificcvfinal\thispagestyle{empty}\fi

%%%%%%%%% ABSTRACT
\begin{abstract}
Open-set image recognition is a challenging topic in computer vision. Most of the existing works in literature focus on learning more discriminative features from the input images, however, they are usually insensitive to either high- or low-frequency components in image features, resulting in a decreasing performance on fine-grained image recognition. To address this problem, we propose a Complementary Frequency-varying Awareness Network that could better capture both high-frequency and low-frequency information, called CFAN. The proposed CFAN consists of three sequential modules: (i) a feature extraction module is introduced for learning preliminary features from the input images; (ii) a frequency-varying filtering module is designed to separate out both high- and low-frequency components from the preliminary features in the frequency domain via a frequency-adjustable filter; (iii) a complementary temporal aggregation module is designed for aggregating the high- and low-frequency components via two Long Short-Term Memory networks into discriminative features. Based on CFAN, we further propose an open-set fine-grained image recognition method, called CFAN-OSFGR, which learns image features via CFAN and classifies them via a linear classifier. Experimental results on 3 fine-grained datasets and 2 coarse-grained datasets demonstrate that CFAN-OSFGR performs significantly better than 9 state-of-the-art methods in most cases.
\end{abstract}

%%%%%%%%% BODY TEXT
%\vspace{-6pt}
\section{Introduction} \label{section: introduction}

Open-set image recognition (OSR) has received more and more attention recently, which aims to both classify known-class images and identify unknown-class images. 
Existing OSR methods \cite{OSR_deep_first, PROSER, CROSR, C2AE, GDFR, RPL, PMAL, OpenHybrid, CGDL, GMVAE-OSR, Capsule, OSR2022_3, GCPL, ARPL, G-openmax, OSRCI, OpenGAN, OSR2022_1, OSR2022_2, OSR_transformer2, MoEP-AE, cite_TCSVT_OSR1, cite_TCSVT_OSR2, cite_TCSVT_OSR3} could be roughly divided into two categories: CNN (Convolutional Neural Network)-based methods \cite{OSR_deep_first, PROSER, CROSR, C2AE, GDFR, RPL, PMAL, OpenHybrid, CGDL, GMVAE-OSR, Capsule, OSR2022_3, GCPL, ARPL, G-openmax, OSRCI, OpenGAN, OSR2022_1, OSR2022_2, cite_TCSVT_OSR1, cite_TCSVT_OSR2, cite_TCSVT_OSR3} and transformer-based methods \cite{OSR_transformer2, MoEP-AE}. Both the two categories of OSR methods have demonstrated their effectiveness to some extent on coarse-grained datasets. However, as shown in \cite{IEEE-Access, good-closed-set}, their performance would decrease in open-set fine-grained image recognition task (OSFGR), which is a sub-task of OSR where the differences among object classes become subtle. Thus in this paper, we focus on OSFGR.

\begin{table}[t]\footnotesize
	\renewcommand\arraystretch{0.5}
	\centering
	\setlength{\abovecaptionskip}{0cm}
	\caption{OSFGR results on high/low-frequency images (HFI/LFI) from Aircraft by ResNet50/SwinB.}
	\begin{tabular}{m{0.9cm}<{\centering}m{0.5cm}<{\centering}m{0.4cm}<{\centering}m{2.2cm}<{\centering}m{2.2cm}<{\centering}}
		\toprule
		Model & Images & ACC & \makecell[c]{AUROC\\ (Easy/Medium/Hard)} & \makecell[c]{OSCR\\ (Easy/Medium/Hard)} \\
		\midrule
		\multirow{2}{*}{ResNet50} & HFI & 0.729 & 0.705/0.668/0.510 & 0.589/0.552/0.537 \\
		& LFI & 0.645 & 0.601/0.524/0.458 & 0.564/0.432/0.378  \\
		\midrule
		\multirow{2}{*}{SwinB} & HFI & 0.794 & 0.789/0.726/0.655 & 0.712/0.681/0.603  \\
		& LFI & 0.836 & 0.818/0.782/0.699 & 0.753/0.722/0.638  \\
%		\midrule
%		ours & HFI & 0.743 & 0.736/0.691/0.608 & 0.624/0.607/0.596   \\
%		(ResNet50) & LFI & 0.718 & 0.685/0.601/0.599 & 0.643/0.562/0.526  \\
%		\midrule
%		ours & HFI & 0.834 & 0.826/0.772/0.723 & 0.755/0.728/0.671   \\
%		(SwinB) & LFI & 0.867 & 0.844/0.812/0.770 & 0.785/0.762/0.689  \\
		\bottomrule
	\end{tabular}
	\label{table: HRI_LRI}
	\vspace{-0.5cm}
\end{table}

As indicated in \cite{fourier1, fourier2, fourier3, fourier4}, the CNN-based closed-set classification methods are generally sensitive to high-frequency information, \emph{e.g.}, object contours in images, while transformer-based closed-set classification methods are generally sensitive to low-frequency information, \emph{e.g.}, flat regions in images. It is also found in the literature that both high-frequency information and low-frequency information have effect on classification, and making better use of high-frequency and low-frequency information would improve the classification performance. Moreover, as done for classifying closed-set coarse-grained images in \cite{fourier2}, we evaluate a typical CNN (ResNet50) and a typical transformer (SwinB) on high-frequency images (called HFI) and low-frequency images (called LFI) under the open-set classification setup, which are obtained from the original images via high-/low-pass filtering in the fine-grained dataset Aircraft \cite{Aircraft}, and the corresponding AUROC and OSCR (the two metrics are defined in Sec. \ref{section: datasets and metrics}) under three difficulty modes (\emph{i.e.}, Easy/Medium/Hard) are reported in Table \ref{table: HRI_LRI}. As seen from this table, ResNet50 performs better on HFI than on LFI, while SwinB performs better on LFI than HFI, demonstrating that they have to be confronted with the same problem in OSFGR as that in the closed-set classification task. 
%Regarding this problem, Table \ref{table: HRI_LRI} reports the OSFGR results on Aircraft of a typical CNN (ResNet50) and a typical transformer (SwinB), both of which are respectively implemented on high-frequency images (called HFI) and low-frequency images (called LFI) obtained from the original images, as done in \cite{fourier2}. These results demonstrate the aforementioned problem in OSFGR. 
Naturally, the following question is raised: ``How to capture both high-frequency and low-frequency information from fine-grained images more effectively for open-set recognition?"

To address this question, we propose a Complementary Frequency-varying Awareness Network (called CFAN) for better capturing both high- and low-frequency information.
CFAN consists of three sequential modules: a feature extraction module, a frequency-varying filtering module, and a complementary temporal aggregation module. The feature extraction module, which could be an arbitrary feature extractor in literature (\emph{e.g.}, ResNet \cite{ResNet}, and SwinB \cite{Swin}), is firstly used to extract the preliminary features from the input fine-grained images. Then, the frequency-varying filtering module is designed to decompose the extracted preliminary features into both high- and low-frequency components in the frequency domain via an explored frequency-adjustable filter. Finally, the complementary temporal aggregation module is explored to aggregate the high- and low-frequency feature components learnt from the frequency-varying filtering module via two LSTMs (Long Short-Term Memory Networks) into discriminative features, inspired by the ability of LSTMs for modeling time-series data in many other visual tasks \cite{LSTM1, LSTM2, LSTM3, LSTM4}. Furthermore, we explore a CFAN-based method to handle the OSFGR task, called CFAN-OSFGR, where the proposed CFAN is used to extract fine-grained image features and then a linear classifier is employed to classify these features. 

The main contributions of this paper are summarized as:

(1) We design a frequency-adjustable filter for decomposing a feature into a set of components with different frequencies. Unlike most of the existing frequency filters \cite{frequency_enhancing1, frequency_enhancing2, frequency_enhancing3} that only extract the component with a specific frequency from the input feature, the designed filter could flexibly switch among a set of filters corresponding to different frequencies. 

(2) We propose a complementary frequency-varying awareness network, CFAN, utilizing the designed frequency-adjustable filter. CFAN could make full use of high-frequency and low-frequency feature components, and could learn more abundant high-frequency and low-frequency information.

(3) We propose the CFAN-OSFGR method for handling the OSFGR task, where the proposed CFAN is concatenated with a linear classifier. Its priority to 9 state-of-the-art methods have been demonstrated in Sec. \ref{section: experiments}.

\section{Related Works} \label{section: related works}

Here, we briefly review the OSR/OSFGR methods in literature, and some typical works that enhance features in the frequency domain.

\subsection{OSR/OSFGR Methods}
%\vspace{-5pt}

In this subsection, we introduce some related works on DNN-based OSR/OSFGR methods. According to different backbones, existing DNN-based OSR/OSFGR methods can be divided into two groups: CNN-based methods which use CNNs as the backbones, and transformer-based methods which use transformers as the backbones.  

\textbf{CNN-based Methods.} Most existing OSR/OSFGR methods are CNN-based methods,
%\cite{OSR_deep_first, PROSER, CROSR, C2AE, GDFR, RPL, PMAL, OpenHybrid, CGDL, GMVAE-OSR, Capsule, OSR2022_3, GCPL, ARPL, G-openmax, OSRCI, OpenGAN, OSR2022_1, OSR2022_2, IEEE-Access, good-closed-set}
most of which are evaluated on coarse-grained datasets. Zhang \emph{et al.} \cite{OpenHybrid} proposed to learn the latent feature representations by jointly using a classifier for classifying known-class features and a density estimator for detecting unknown-class features. They adopted a resflow-net \cite{resflow} to model the known-class likelihood scores, and simultaneously used a classifier in the latent feature space for classification. Kong and Ramanan \cite{OpenGAN} proposed to use fake data generated by a GAN as unknown-class data to augment the training set. It was found that, even without data generation, building a discriminator over the known-class latent feature representations was also effective for open-set recognition, where the discriminator was then used for distinguishing known-class features from unknown-class ones in inference. They used a VGG \cite{VGG} as the feature extractor. Chen \emph{et al.} \cite{ARPL} mined unknown-class features in the extra-class space of each known class via a ResNet \cite{ResNet}, then trained the model with known-class samples and these features adversarially for encouraging the known-class feature space to be more compact. Yang \emph{et al.} \cite{GCPL} modeled the feature distribution of each known class as a Gaussian mixture for learning more discriminative features via a ResNet by prototype learning, such that unknown-class features could be detected since their probabilities of belonging to these known-class distributions would be low. Similarly, Cao \emph{et al.} \cite{GMVAE-OSR} also modeled known-class features as multiple mixtures of Gaussian, but they used a VGG-based Gaussian Mixture Variational Auto-Encoders (GMVAE) \cite{GMVAE} for modeling distributions in the latent feature space. Besides, a few OSFGR methods \cite{IEEE-Access, good-closed-set} have been proposed, aiming to learn more discriminative features. Dai \emph{et al.} \cite{IEEE-Access} analyzed the characteristics of different classification scores and chose the class activation mapping values outputted from a VGG for preserving fine-grained information. Vaze \emph{et al.} \cite{good-closed-set} proposed to take full advantage of multiple training strategies for improving the discriminability of known-class features extracted from a ResNet backbone.

\begin{figure*}[t]
	\begin{center}
		\setlength{\abovecaptionskip}{0.cm}
		\includegraphics[height=7.1cm,width=17.3cm]{{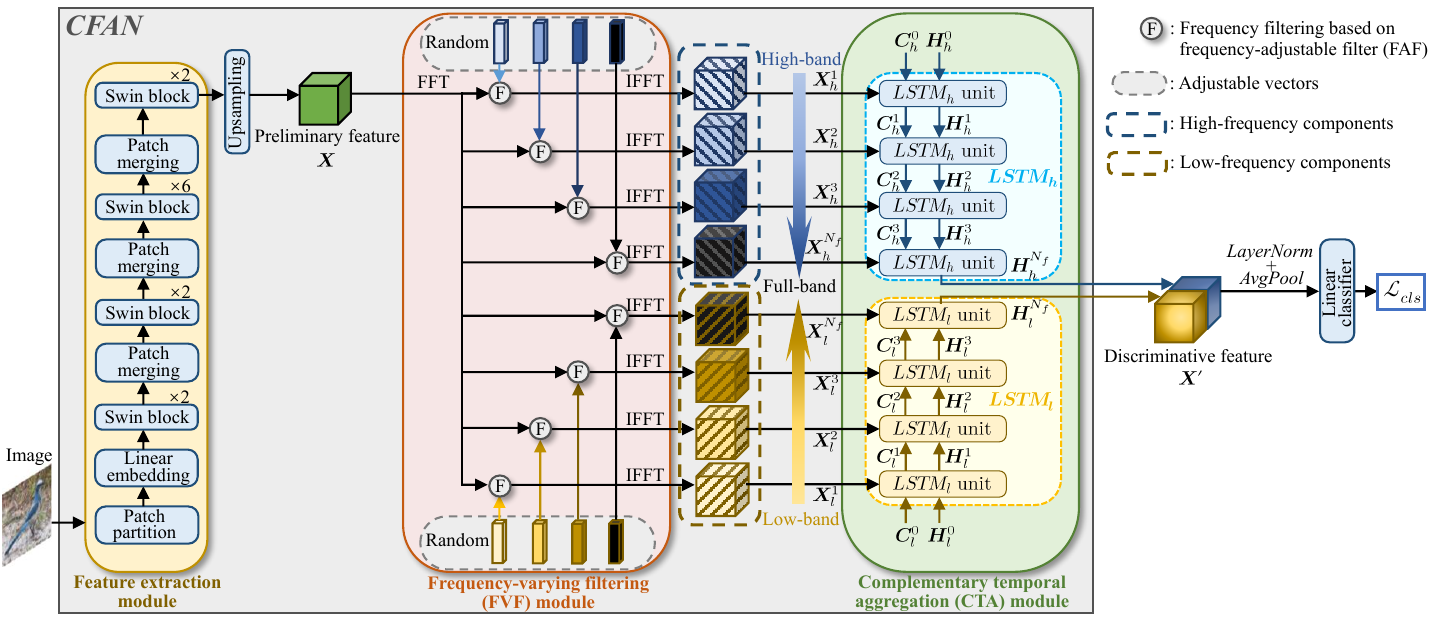}}
	\end{center}
	\vspace{-0.4cm}
	\caption{Architecture of the proposed CFAN-OSFGR method: Firstly, an image is fed into the proposed CFAN for obtaining a corresponding discriminative feature $\boldsymbol{X}'$, which consists of three sequential modules, including a feature extraction module (ResNet50 \cite{ResNet} and SwinB \cite{Swin} are respectively used here, and we show the SwinB as an example) for extracting a preliminary image feature $\boldsymbol{X}$, a frequency-varying filtering module for separating out the high- and the low-frequency components from the preliminary feature in the frequency domain, and a complementary temporal aggregation module for aggregating the high- and the low-frequency feature components into a discriminative feature. Then, the discriminative feature is fed into a linear classifier. CFAN-OSFGR is trained by minimizing the classification loss $\mathcal{L}_{cls}$.}
	\label{fig: architecture}
	\vspace{-0.3cm}
\end{figure*}

\textbf{Transformer-based Methods.} Recently, vision transformers have obtained more and more attention in many vision tasks, due to their strong ability to learn discriminative features based on multiple self-attention operations. Inspired by the success of vision transformers in closed-set image recognition tasks \cite{ViT, TIT, PVT, Swin}, a few transformer-based methods \cite{OSR_transformer2, MoEP-AE} have been proposed. Sun \emph{et al.} \cite{MoEP-AE} built a transformer-based encoder for encoding known-class images into the distribution parameters about mixtures of exponential power distributions in a latent space, and took the features sampled from the learnt distributions for decoding. A classifier was used for classifying the encoded distribution parameters. Azizmalayeri and Rohban \cite{OSR_transformer2} also adopted a transformer as the backbone for learning discriminative feature representations, and they integrated various data augmentation strategies for further boosting the feature discriminability.

However, as mentioned in Sec. \ref{section: introduction}, the above CNN-based or transformer-based methods are usually insensitive to either high-frequency or low-frequency components in image features. Hence, we aim to design the complementary frequency-varying awareness network for capture both high-frequency and low-frequency information more effectively for open-set recognition.

%\vspace{-5pt}
\subsection{Frequency Based Feature Enhancement}
%\vspace{-5pt}

%It is noted that a good feature should \hl{represent both high- and low-frequency components}, as indicated in \cite{add1, add2, add3}. However, the CNN backbones in the aforementioned CNN-based methods have a deficiency in capturing high-frequency signals as indicated in \cite{fourier2, fourier3, fourier4}, while the transformer backbones in the aforementioned transformer-based methods cannot well capture low-frequency signals as indicated in \cite{fourier1}, limiting the discriminability of the features extracted from these backbones to some extent. This limitation motivates us to design the CFAN-OSFGR method for \hl{complementing the high- and low-frequency information}, which will be described in detail in the following section.

Recently, some frequency-based feature enhancement works have been proposed in other visual task \cite{frequency_enhancing1, frequency_enhancing2, frequency_enhancing3}, which also aim to make better use of high-frequency and low-frequency information. Rao \emph{et al.} \cite{frequency_enhancing1} proposed a global filter network for learning long-term spatial dependencies in an image, they used learnable filters at different layers and encouraged each filter to pass frequency component at an appropriate band. Liu \emph{et al.} \cite{frequency_enhancing2} proposed a global spectral filter memory network, aiming to learn long-term spatial dependencies between different video frames, and they used the traditional Gaussian filter for frequency filtering. Qin \emph{et al.} \cite{frequency_enhancing3} proposed a multi-spectral channel attention, which used the conventional global average pooling operation instead of frequency filters for feature decomposition in the frequency domain. 

Different from these methods, the proposed method uses an adjustable frequency filter which can obtain a set of high- and low-frequency components at various frequency bands by adjusting the adjustable vectors, aiming to make use of more abundant frequency information.

\section{Complementary Frequency-Varying Awareness Network for OSFGR} \label{section: methods}

\subsection{Complementary Frequency-Varying Awareness Network}  \label{section: network}

Here, we propose the Complementary Frequency-varying Awareness Network (CFAN), consisting of a feature extraction module, a frequency-varying filtering module, and a complementary temporal aggregation module. Firstly, we introduce the whole architecture of CFAN and the feature extraction module. Then, we describe the other two modules in the proposed CFAN in detail.

\noindent \normalsize{\textbf{3.1.1 Architecture and Feature Extraction Module}}

As seen from Fig. \ref{fig: architecture}, CFAN takes object images as its inputs, and aims to output discriminative features. It contains three sequential modules, a feature extraction module, a frequency-varying filtering module, and a complementary temporal aggregation module. The feature extraction module is firstly used to learn preliminary features from the input images. Once the preliminary features have been learnt from the feature extraction module, the frequency-varying filtering module is used for converting the preliminary features into time-series features that cover various high- and low-frequency bands. Finally, the complementary temporal aggregation module is used for aggregating the high- and low-frequency components into discriminative features. 

It has to be pointed out that many feature extractors in literature (\emph{e.g.}, VGG \cite{VGG}, ResNet \cite{ResNet}, SwinB \cite{Swin}, etc.) could be straightforwardly used as the feature extraction module. Here, we simply use ResNet50 \cite{ResNet} and SwinB \cite{Swin} (as shown in the left-most yellow box in Fig. \ref{fig: architecture}) as the feature extraction module, respectively. The frequency-varying filtering module and the complementary temporal aggregation module would be described in detail in the following subsections.

\vspace{0.1cm}
\noindent \normalsize{\textbf{3.1.2 Frequency-Varying Filtering Module}}
\vspace{0.1cm}

\begin{figure}[t]
	\begin{center}
		\setlength{\abovecaptionskip}{0.cm}
		\includegraphics[height=2.4cm,width=7.5cm]{{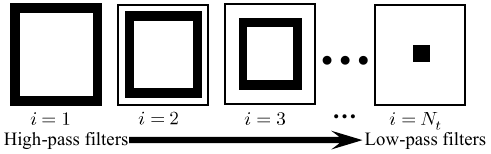}}
	\end{center}
	\vspace{-0.4cm}
	\caption{Sketch of the sequence of band-pass template filters $\{ \mathbf{T}_i \}_{i=1}^{N_t}$ at each channel in the frequency-varying filtering module. Each value in these filters is either 1 (in black) or 0 (in white).}
	\label{fig: templates}
	\vspace{-1cm}
\end{figure}

The frequency-varying filtering module is designed for separating out both high- and low-frequency components from the preliminary features in the frequency domain via an explored frequency-adjustable filter. As shown in the red box in Fig. \ref{fig: architecture}, this module takes the preliminary feature extracted from each input image by the feature extraction module as its input, and outputs two time series of feature components at various high and low frequencies respectively.

\textbf{Frequency-Adjustable Filter.} In order to extract various bands of high- and low-frequency information flexibly, we design the frequency-adjustable filter, which is a weighted combination of a sequence of $N_t$ (here we set $N_t =20$) band-pass template filters $\{ \mathbf{T}_i \}_{i=1}^{N_t}$ as shown in Fig. \ref{fig: templates}. This filter has a high-pass form $\boldsymbol{F}_h$ and a low-pass form $\boldsymbol{F}_l$ as:
%\vspace{-0.4cm}
\begin{align}
\boldsymbol{F}_h = \sum_{i=1}^{N_t} \boldsymbol{f}^h_i \mathbf{T}_i , \ \
\boldsymbol{F}_l = \sum_{i=1}^{N_t} \boldsymbol{f}^l_i \mathbf{T}_i  \label{4}
\end{align}
where $\{ \boldsymbol{f}^h_i \}_{i=1}^{N_t}$ (also $\{ \boldsymbol{f}^l_i \}_{i=1}^{N_t}$) is a set of weighting coefficients. Here, we use the exponential power (EP) function values to assign these coefficients, considering that the EP function is a member of the function family whose function shape can be flexibly adjusted as indicated in \cite{EP_function1, EP_function2, EP_function3}. According to the parametric form suggested in \cite{EP_function1}, the complete form of the EP function is formulated as:
%\vspace{-0.3cm}
\begin{align}
EP(\boldsymbol{m}) = \dfrac{1}{2 \boldsymbol{\sigma} \boldsymbol{p}^{\frac{1}{\boldsymbol{p}}} \Gamma ( 1+\frac{1}{\boldsymbol{p}} )} e^{-\frac{| \boldsymbol{m} - \boldsymbol{\mu} |^{\boldsymbol{p}}}{\boldsymbol{p} \boldsymbol{\sigma}^{\boldsymbol{p}}}}
\end{align}
where $\boldsymbol{m}$ is the independent variable, $\Gamma(\cdot)$ is the Gamma function, and $\{ \boldsymbol{\mu}, \boldsymbol{\sigma}, \boldsymbol{p} \} \ (\boldsymbol{\sigma}>0, \boldsymbol{p}>0)$ is a group of parameters that control the position, scale, and shape of the EP function respectively. It is noted that the shape parameter $\boldsymbol{p}$ plays a leading role in controlling the shape of the EP function, hence, we use $\boldsymbol{p}$ as an adjustable vector to switch the shape of the EP function from steep to gentle, which switches the frequency filter whose coefficients are assigned by the EP function values from a high- or low-pass filter to a full-pass filter. In order to limit the function values within $(0,1)$, we discard the coefficient part before the exponential power part of this formula. It is noted that the EP function is axial-symmetrical to the axis $\boldsymbol{m} = \boldsymbol{\mu}$, hence, the sequence of weighting coefficients $\{ \boldsymbol{f}^h_i \}_{i=1}^{N_t}$ for the high-pass filter $\boldsymbol{F}_h$ and the sequence of weighting coefficients $\{ \boldsymbol{f}^l_i \}_{i=1}^{N_t}$ for the low-pass filter $\boldsymbol{F}_l$ can be obtained by simply setting the axis of symmetry at $\boldsymbol{m} = \boldsymbol{0} $ and $\boldsymbol{m} = N_t \cdot \boldsymbol{I} $ (where $\boldsymbol{I}$ is a vector with all values being $1$) respectively and taking the EP function values corresponding to the evenly-spaced independent variable values which are sampled over the interval $(\boldsymbol{0}$, $N_t \cdot \boldsymbol{I})$ (here we sample $\boldsymbol{m}=0.5 \cdot \boldsymbol{I}, 1.5 \cdot \boldsymbol{I}, ..., (N_t - 0.5) \cdot \boldsymbol{I}$) as the weighting coefficients for the template filters. Thus, the $i$-th elements ($i \in \{ 1,2,..., N_t \}$) $\boldsymbol{f}^h_i$ and $\boldsymbol{f}^l_i$ in the two sequences $\{ \boldsymbol{f}^h_i \}_{i=1}^{N_t}$ and $\{ \boldsymbol{f}^l_i \}_{i=1}^{N_t}$ can be formulated as:
%\vspace{-0.2cm}
\begin{align}
&\boldsymbol{f}^h_i  = e^{-\frac{| \boldsymbol{m}- \boldsymbol{0}|^{\boldsymbol{p}_h}}{\boldsymbol{p}_h \boldsymbol{\sigma}^{\boldsymbol{p}_h}}}  , \ \ 
\boldsymbol{f}^l_i  = e^{-\frac{| \boldsymbol{m}- N_t \cdot \boldsymbol{I}|^{\boldsymbol{p}_l}}{\boldsymbol{p}_l \boldsymbol{\sigma}^{\boldsymbol{p}_l}}}  \nonumber \\
\mathrm{s.t.} \ \ \ \ \ &\boldsymbol{m}= (i-0.5) \cdot \boldsymbol{I}, \ \ 
\boldsymbol{\sigma} = N_t \cdot \boldsymbol{I}
\end{align}
%\begin{align}
%&\{ \boldsymbol{f}^h_i \}_{i=1}^{N_t}  = e^{-\frac{| \boldsymbol{m}- \boldsymbol{0}|^{\boldsymbol{p}_h}}{\boldsymbol{p}_h \boldsymbol{\sigma}^{\boldsymbol{p}_h}}} , \ \ \ \ 
%\{ \boldsymbol{f}^l_i \}_{i=1}^{N_t} = e^{-\frac{| \boldsymbol{m}- N_t \cdot \boldsymbol{I}|^{\boldsymbol{p}_l}}{\boldsymbol{p}_l \boldsymbol{\sigma}^{\boldsymbol{p}_l}}}  \nonumber \\
%\mathrm{s.t.} \ \ &\boldsymbol{m}=0.5 \cdot \boldsymbol{I},\ 1.5 \cdot \boldsymbol{I}, \ ..., \ (N_t - 0.5) \cdot \boldsymbol{I}
%\end{align}
where the values in the two adjustable vectors $\boldsymbol{p}_h$ and $\boldsymbol{p}_l$ are separately adjusted varying in $[p_{min},p_{max}]$, where $p_{min}$ and $p_{max}$ are two preset constants. The high-pass filter $\boldsymbol{F}_h$ (or the low-pass filter $\boldsymbol{F}_l$) inclines to high-pass filtering (or low-pass filtering) when the values in $\boldsymbol{p}_h$ (or $\boldsymbol{p}_l$) get closer to $p_{min}$, and inclines to full-pass filtering when the values get closer to $p_{max}$.

\textbf{Frequency Filtering Process.} Here, we describe the filtering process in the frequency-varying filtering module. Firstly, we conduct the Fast Fourier Transform (FFT) at each channel on the preliminary feature map $\boldsymbol{X}$ with $N_c$ channels: $\boldsymbol{Z} = \mathcal{F}(\boldsymbol{X})$, and centralize the complex spectrum $\boldsymbol{Z}$. Next, a time series of high-frequency components $\{ \boldsymbol{Z}_h^t \}_{t=1}^{N_f}$ and a time series of low-frequency components $\{ {\boldsymbol{Z}_l^t} \}_{t=1}^{N_f}$ can be obtained from $\boldsymbol{Z}$ by a time series of high-pass filters $\{ {\boldsymbol{F}_h^t} \}_{t=1}^{N_f}$ and a time series of low-pass filters $\{ {\boldsymbol{F}_l^t} \}_{t=1}^{N_f}$ by utilizing the designed frequency-adjustable filter, whose $t$-th elements ($t \in \{ 1,2,...,N_f \}$) can be formulated as:
%\vspace{-0.1cm}
\begin{align}
\boldsymbol{Z}_h^t = \boldsymbol{Z} \odot \boldsymbol{F}_h^t ,  \ \ 
\boldsymbol{Z}_l^t = \boldsymbol{Z} \odot \boldsymbol{F}_l^t   \label{5}
\end{align}
%\begin{align}
%\{ \boldsymbol{Z}_h^t = \boldsymbol{Z} \odot \boldsymbol{F}_h^t \}_{t=1}^{N_f}, \ \ \ \ 
%\{ \boldsymbol{Z}_l^t = \boldsymbol{Z} \odot \boldsymbol{F}_l^t \}_{t=1}^{N_f}   \label{5}
%\end{align}
where `$\odot$' represents the element-wise product operator; $N_f$ is the length of the time series. $\{ {\boldsymbol{F}_h^t} \}_{t=1}^{N_f}$ and $\{ {\boldsymbol{F}_l^t} \}_{t=1}^{N_f}$ are obtained by two time series of adjustable vectors $\{ \boldsymbol{p}_h^t \}_{t=1}^{N_f}$ and $\{ \boldsymbol{p}_l^t \}_{t=1}^{N_f}$ whose elements are evenly-spaced sampled over the intervals $[ \boldsymbol{p}_h^1, p_{max} \cdot \boldsymbol{I} ]$ and $[ \boldsymbol{p}_l^1, p_{max} \cdot \boldsymbol{I} ]$ respectively (where the values in the initial adjustable vectors $\boldsymbol{p}_l^1$ and $\boldsymbol{p}_h^1$ are randomly sampled from $U(0,1)$). Then, the two time series of feature components are decentralized and transformed by the Inverse Fast Fourier Transform (IFFT): $\{ \boldsymbol{X}_h^t = \mathcal{F}^{-1}(\boldsymbol{Z}_h^t) \}_{t=1}^{N_f}, \ \  \{ \boldsymbol{X}_l^t = \mathcal{F}^{-1}(\boldsymbol{Z}_l^t) \}_{t=1}^{N_f} $. Thus, we obtain two time series of feature components $\{ \boldsymbol{X}_h^t \}_{t=1}^{N_f}$ and $\{ \boldsymbol{X}_l^t \}_{t=1}^{N_f}$ at various high- and low-frequency bands respectively.

\vspace{0.1cm}
\noindent \normalsize{\textbf{3.1.3 Complementary Temporal Aggregation Module}}
\vspace{0.1cm}

The complementary temporal aggregation module is designed to aggregate the time series of high-frequency components and the time series of low-frequency components obtained from the frequency-varying filtering module, and output a discriminative feature. Considering that LSTMs have shown the superiority in modeling time-series data for handling many other visual tasks \cite{LSTM_other1, LSTM_other2, LSTM_other3, LSTM_other4, LSTM_other5}, we use two LSTMs to respectively model the temporal dependences of high-frequency components and low-frequency components in this module, as shown in the green box of Fig. \ref{fig: architecture}.
 
Specifically, let $\{ \boldsymbol{X}_h^t \}_{t=1}^{N_f}$ represent the high-frequency components outputted from the frequency-varying filtering module, and $\{ \boldsymbol{X}_l^t \}_{t=1}^{N_f}$ represent the low-frequency components. At the $t$-th moment ($t \in \{1,2,...,N_f \}$), the Hidden and Cell states are updated by:
%\vspace{-0.2cm}
\begin{align}
\boldsymbol{H}_h^t &= {LSTM_h}(\boldsymbol{H}_h^{t-1}, \boldsymbol{C}_h^{t-1}, \boldsymbol{X}_h^t) , \\
\boldsymbol{H}_l^t &= LSTM_l(\boldsymbol{H}_l^{t-1}, \boldsymbol{C}_l^{t-1}, \boldsymbol{X}_l^t)   \label{8}
\end{align}
where $\boldsymbol{H}_h^t$ and $\boldsymbol{H}_l^t$ represent the Hidden states, $\boldsymbol{C}_h^t$ and $\boldsymbol{C}_l^t$ represent the Cell states at the $t$-th moment; $\boldsymbol{X}_h^t$ and $\boldsymbol{X}_l^t$ represent the $t$-th element of the time series $\{ \boldsymbol{X}_h^t \}_{t=1}^{N_f}$ and $\{ \boldsymbol{X}_l^t \}_{t=1}^{N_f}$, respectively. Finally, a discriminative feature $\boldsymbol{X}'$ can be obtained by concatenating the two updated Hidden states at the $N_f$-th moment along the channel dimension: $\boldsymbol{X}' = [\boldsymbol{H}_h^{N_f}; \boldsymbol{H}_l^{N_f}]$.

\subsection{CFAN-OSFGR}  \label{section: training and testing}

Here, we introduce the CFAN-OSFGR method for handling the OSFGR task. The CFAN is firstly integrated with a LayerNorm layer for normalization, an average pooling layer for dimension reduction, and a linear classifier for recognition, as shown in Fig. \ref{fig: architecture}. Then, the training strategy and the inference strategy are described as follows.

\textbf{Training.} The model is trained with a cross-entropy classification loss:
%\vspace{-0.2cm}
\begin{align}
\mathcal{L}_{cls} = - \frac{1}{N_b} \sum_{k=1}^{N_b} \mathrm{log} \, p_k  \label{11}
\end{align}
where $N_b$ represents the batch size, and $p_k$ represents the probability of the $k$-th image in the current batch corresponding to the ground-truth class. 

\textbf{Inference.} The score outputted from the classifier of the proposed model is used for inference, which is defined as:
%\vspace{-0.2cm}
\begin{align}
s = \max_{c} \{ y_c \} , \ \ c \in \{ 1,2,...,C \}  \label{12}
\end{align}
where $C$ is the number of the known classes, $y_c$ indicates the $c$-th ($c \in \{ 1,2,...,C \}$) element of the logit vector outputted from the classifier corresponding to the $c$-th class. Besides, a threshold $\theta$, which is chosen to make 90\% validation images be correctly recognized as known classes, is used for classifying known-class images and identifying unknown-class images by comparing with the score $s$:
%\vspace{-0.2cm}
\begin{align}
\mathrm{prediction}= \left\{
\begin{aligned}
&\mathrm{argmax}_{c \in \{ 1,2,...,C \}} \, y_c , \ \ \ \ \  \mathrm{if} \ s \geq \theta \\
&\mathrm{unknown \; \; classes} , \ \ \ \ \ \ \ \ \ \ \, \,  \mathrm{if} \ s < \theta
\end{aligned}  \label{13}
\right.   
\end{align}

\section{Experiments}  \label{section: experiments}

\subsection{Datasets and Metrics}  \label{section: datasets and metrics}

\textbf{Datasets.} The proposed CFAN-OSFGR method is evaluated on 3 fine-grained datasets (including Aircraft \cite{Aircraft}, CUB \cite{CUB}, and Stanford-Cars \cite{Stanford-Cars}) and 2 coarse-grained datasets which are relatively difficult in the OSR task (including CIFAR+10/+50 \cite{CIFAR10, CIFAR100} and TinyImageNet \cite{TinyImageNet}) under two dataset settings:

(1) \textit{Standard-Dataset Setting}. Under this setting, the known-class and unknown-class images are from the same dataset. Aircraft \cite{Aircraft} contains 100-class aircraft images with attributes, 50 classes of which are selected as the known classes. The rest classes are further divided into three modes: `Easy', `Medium', and `Hard' according to their attribute similarity to the known classes, and we follow the 20/17/13 splitting manner for splitting the unknown classes as done in \cite{good-closed-set}. CUB \cite{CUB} contains 200-class bird images with attributes, 100 classes of which are selected as the known classes, and we follow the 32/34/34 splitting manner for splitting the unknown classes as done in \cite{good-closed-set}.
%CUB \cite{CUB} contains 200-class bird images with attributes, 100 classes of which are selected as the known classes. The rest classes are further divided into three modes: `Easy', `Medium', and `Hard' according to their attribute similarity to the known classes, and we follow the 32/34/34 splitting manner for splitting the unknown classes as done in \cite{good-closed-set}. Aircraft \cite{Aircraft} contains 100-class aircraft images with attributes, 50 classes of which are selected as the known classes, and we follow the 20/17/13 splitting manner for splitting the unknown classes as done in \cite{good-closed-set}. 
Stanford-Cars \cite{Stanford-Cars} contains 196-class car images, the first 98 classes of which are selected as the known classes while the rest 98 classes are used as the unknown classes. In CIFAR+10/+50, 10 classes in CIFAR10 \cite{CIFAR10} are used as the known classes, while 10 or 50 non-overlapping classes in CIFAR100 \cite{CIFAR100} are used as the unknown classes. TinyImageNet \cite{TinyImageNet} contains 200-class natural images, 20 classes of which are used as the known classes, while the rest 180 classes are used as the unknown classes.   

(2) \textit{Cross-Dataset Setting}. This setting is configured by using the 50 split known classes in Aircraft as the known classes while using all of the 200- and 196-class testing images in CUB and Stanford-Cars as the unknown-class images, respectively.

\textbf{Metrics.} The following evaluation metrics are used under the above dataset settings:

(1) \textit{Standard-Dataset Setting}. On the coarse-grained datasets, we use two metrics for evaluation as done in \cite{MoEP-AE, OpenHybrid, OSR_transformer2, ARPL, OpenGAN}: (i) AUROC which measures the open-set detection performance by regarding the OSFGR task as a binary classification task (\emph{i.e.}, classification between known classes and unknown classes), and (ii) ACC (\emph{i.e.}, the top-1 accuracy that is widely used in the closed-set classification task) which measures the closed-set classification performance. On the fine-grained datasets, in addition to AUROC and ACC, we also use OSCR (\emph{i.e.}, the open-set classification rate \cite{OSCR}), which is a threshold-independent metric that simultaneously measures the open-set detection performance and the closed-set classification performance, as done in \cite{good-closed-set}.

(2) \textit{Cross-Dataset Setting}. As done in \cite{IEEE-Access, MoEP-AE, OpenGAN}, we use macro-F1 score which is a threshold-dependent metric that measures the open-set classification performance by taking the unknown classes as the ($C+1$)-th class.

\begin{table*}\footnotesize
	\renewcommand\arraystretch{0.5}
	\centering
	\setlength{\abovecaptionskip}{0cm}  %段前
	\setlength{\belowcaptionskip}{-0.2cm} %段后
	\caption{Comparison of ACC, AUROC, and OSCR results under the \textit{standard-dataset setting} on Aircraft.}
	\begin{tabular}{m{2.2cm}<{\centering}m{2.8cm}<{\raggedright}m{1.5cm}<{\centering}m{3.8cm}<{\centering}m{3.8cm}<{\centering}}
		\toprule
		Backbone & Method & ACC & \makecell[c]{AUROC\\ (Easy/Medium/Hard)} & \makecell[c]{OSCR\\ (Easy/Medium/Hard)} \\ 
		\midrule
		\multirow{8}{*}{CNN} & OpenHybrid \cite{OpenHybrid} & 0.735 & 0.884/0.844/0.780 & 0.718/0.687/0.629   \\
		& OpenGAN \cite{OpenGAN} & 0.807 & 0.841/0.821/0.688 & 0.795/0.754/0.651  \\
		& ARPL \cite{ARPL} & 0.917 & 0.867/0.852/0.674 & 0.814/0.820/0.657 \\
		& GCPL \cite{GCPL} & 0.823 & 0.837/0.828/0.709 & 0.805/0.761/0.652 \\
		& GMVAE-OSR \cite{GMVAE-OSR} & 0.814 & 0.849/0.833/0.687 & 0.799/0.753/0.661 \\
		& CAMV \cite{IEEE-Access} & 0.868 & 0.880/0.844/0.730 & 0.801/0.772/0.680 \\
		& Cross-Entropy+ \cite{good-closed-set} & 0.917 & 0.907/0.864/0.776 & 0.868/0.831/0.754  \\
		\cmidrule{2-5}
		& CFAN-OSFGR & \textbf{0.918} & \textbf{0.922/0.873/0.784} & \textbf{0.873/0.847/0.771}  \\
		\midrule
		\multirow{9}{*}{SwinB} & Backbone & 0.896 & 0.848/0.827/0.723 & 0.799/0.778/0.686   \\
		& OpenHybrid & 0.901 & 0.889/0.830/0.733 & 0.823/0.781/0.689  \\
		& ARPL & 0.915 & 0.880/0.833/0.720 & 0.836/0.793/0.691   \\
		& Cross-Entropy+ & 0.903 & 0.859/0.841/0.712 & 0.814/0.796/0.681  \\
		& Trans-AUG \cite{OSR_transformer2} & 0.898 & 0.873/0.842/0.739 & 0.814/0.787/0.689  \\
		& MoEP-AE-OSR \cite{MoEP-AE} & 0.894 & 0.907/0.876/0.752 & 0.831/0.805/0.701   \\
		\cmidrule{2-5}
		& GFNet \cite{frequency_enhancing1} & 0.901 & 0.871/0.849/0.755 & 0.817/0.796/0.743  \\
		& FcaNet \cite{frequency_enhancing3} & 0.903 & 0.863/0.856/0.748 & 0.815/0.807/0.735  \\
		\cmidrule{2-5}
		& CFAN-OSFGR & \textbf{0.917} & \textbf{0.913/0.892/0.839} & \textbf{0.864/0.845/0.803}   \\
		\bottomrule
	\end{tabular}
	\label{table: Aircraft}
	\vspace{-0.1cm}
\end{table*}

\begin{table*}\footnotesize
	\renewcommand\arraystretch{0.5}
	\centering
	\setlength{\abovecaptionskip}{0cm}  %段前
	\setlength{\belowcaptionskip}{-0.2cm} %段后
	\caption{Comparison of ACC, AUROC, and OSCR results under the \textit{standard-dataset setting} on CUB.}
	\begin{tabular}{m{2.2cm}<{\centering}m{2.8cm}<{\raggedright}m{1.5cm}<{\centering}m{3.8cm}<{\centering}m{3.8cm}<{\centering}}
		\toprule
		Backbone & Method & ACC & \makecell[c]{AUROC\\ (Easy/Medium/Hard)} & \makecell[c]{OSCR\\ (Easy/Medium/Hard)}  \\
		\midrule
		\multirow{8}{*}{CNN} & OpenHybrid \cite{OpenHybrid} & 0.683 & 0.873/0.862/0.739 & 0.662/0.649/0.531   \\
		& OpenGAN \cite{OpenGAN} & 0.799 & 0.801/0.765/0.707 & 0.725/0.706/0.648   \\		
		& ARPL \cite{ARPL} & \textbf{0.863} & 0.814/0.772/0.703 & 0.747/0.710/0.659  \\
		& GCPL \cite{GCPL} & 0.783 & 0.805/0.732/0.645 & 0.711/0.619/0.565   \\
		& GMVAE-OSR \cite{GMVAE-OSR} & 0.725 & 0.826/0.750/0.704 & 0.694/0.628/0.553   \\
		& CAMV \cite{IEEE-Access} & 0.836 & 0.845/0.802/0.709 & 0.746/0.715/0.640   \\
		& Cross-Entropy+ \cite{good-closed-set} & 0.862 & \textbf{0.883}/0.823/0.763 & 0.798/0.754/0.708   \\
		\cmidrule{2-5}
		& CFAN-OSFGR & \textbf{0.863} & \textbf{0.883/0.871/0.775} & \textbf{0.803/0.762/0.716}   \\
		\midrule
		\multirow{9}{*}{SwinB} & Backbone & 0.949 & 0.945/0.875/0.804 & 0.908/0.848/0.781   \\
		& OpenHybrid & 0.950 & 0.953/0.881/0.808 & 0.918/0.855/0.783   \\
		& ARPL & 0.952 & 0.948/0.877/0.810 & 0.912/0.850/0.788   \\
		& Cross-Entropy+ & \textbf{0.953} & 0.950/0.879/0.815 & 0.917/0.854/0.794   \\
		& Trans-AUG \cite{OSR_transformer2} & 0.950 & 0.953/0.882/0.818 & 0.914/0.854/0.795   \\
		& MoEP-AE-OSR \cite{MoEP-AE} & 0.948 & 0.957/0.889/0.814 & 0.915/0.856/0.787  \\
		\cmidrule{2-5}
%		& CFAN-OSFGR & 0.947 & 0.954/\textbf{0.909/0.833} & 0.915/\textbf{0.877/0.810}  \\
		& GFNet \cite{frequency_enhancing1} & 0.949 & 0.951/0.882/0.812 & 0.913/0.851/0.790  \\
		& FcaNet \cite{frequency_enhancing3} & 0.949 & 0.948/0.887/0.809 & 0.911/0.858/0.787  \\
		\cmidrule{2-5}
		& CFAN-OSFGR & 0.950 & \textbf{0.959/0.899/0.828} & \textbf{0.921/0.870/0.806}  \\
		\bottomrule
	\end{tabular}
	\label{table: CUB}
	\vspace{-0.4cm}
\end{table*}

\subsection{Implementation Details}  \label{section: details}

The images from Aircraft and CUB are resized to $448 \times 448$ as done in \cite{good-closed-set}, while those from Stanford-Cars are resized to $224 \times 224$. The ResNet50 and the `base' version of Swin Transformer (\emph{i.e.}, SwinB \cite{Swin}) are used as the feature extractor respectively in the feature extraction module.
%, and its pretrained weights based on the ImageNet-22K dataset \cite{ImageNet-22K} is used for the model initialization. 
We use an SGD optimizer with the learning rate of $3 \times 10^{-4}$ and the weight decay of $1 \times 10^{-4}$, as well as an AdamW optimizer \cite{AdamW} with the learning rate of $5 \times 10^{-5}$ and the weight decay of $0.01$ for optimizing the CNN part and the transformer part, respectively. $N_t$, $N_f$, $N_b$ and $N_c$ are set to $20$, $4$, $32$, and $128$, respectively; $p_{min}$ is set to $0.2$ for avoiding filter values being too small, and $p_{max}$ is set to $20$ for avoiding the numerical overflow. The originally extracted features are $4$-times upsampled for obtaining the preliminary features, and the channels are simultaneously $8$-times pruned for decreasing the model complexity. At the inference stage, the values in $\boldsymbol{p}_h^1$ and $\boldsymbol{p}_l^1$ are simply set to $1$, since the model has been trained to be robust to the variance of these values.

\subsection{Evaluation}   \label{section: evalution}

\vspace{0.1cm}
\noindent \normalsize{\textbf{4.3.1 Evaluation on the Fine-Grained Datasets}}
\vspace{0.1cm}

\textbf{Evaluation Under the \textit{Standard-Dataset Setting}.} Considering that only a few works (CAMV \cite{IEEE-Access} and Cross-Entropy+ \cite{good-closed-set}) are specially designed for handling the OSFGR task, we also compare the proposed CFAN-OSFGR method with 7 state-of-the-art OSR methods (OpenHybrid \cite{OpenHybrid}, OpenGAN \cite{OpenGAN}, ARPL \cite{ARPL}, GCPL \cite{GCPL}, GMVAE-OSR \cite{GMVAE-OSR}, MoEP-AE-OSR \cite{MoEP-AE} and Trans-AUG \cite{OSR_transformer2}). In addition, considering the SwinB-based methods are few, we evaluate one relatively better OSFGR method (Cross-Entropy+) and two relatively better OSR methods (OpenHybrid and ARPL) by replacing their original backbones with SwinB for further comparison. Besides, in order to evaluate the effectiveness of the proposed frequency-based feature enhancement open-set recognition method, we also modify two frequency-based feature enhancement closed-set classification methods (GFNet \cite{frequency_enhancing1} and FcaNet \cite{frequency_enhancing3} mentioned in Sec. \ref{section: related works}), whose backbones are replaced with SwinB.
%Besides, to eliminate the influence of different backbones, the CNN backbones of the comparative methods are unified as ResNet50 or SwinB.

\begin{table}[t]\footnotesize
	\renewcommand\arraystretch{0.5}
	\centering
	\setlength{\abovecaptionskip}{0cm}
	\caption{Comparison of ACC, AUROC, and OSCR results under the \textit{standard-dataset setting} on Stanford-Cars.}
	\begin{tabular}{m{0.9cm}<{\centering}m{2.7cm}<{\raggedright}m{0.9cm}<{\centering}m{0.8cm}<{\centering}m{0.8cm}<{\centering}}
		\toprule
		Backbone & Method & ACC & AUROC & OSCR  \\
		\midrule
		\multirow{7}{*}{CNN} & OpenHybrid \cite{OpenHybrid} & 0.675 & 0.699 & 0.637 \\
		& OpenGAN \cite{OpenGAN} & 0.704 & 0.763 & 0.682 \\	
		& ARPL \cite{ARPL} & 0.752 & 0.839 & 0.726 \\
		& GCPL \cite{GCPL} & 0.688 & 0.735 & 0.663   \\
		& GMVAE-OSR \cite{GMVAE-OSR} & 0.683 & 0.756 & 0.670   \\
		& CAMV \cite{IEEE-Access} & 0.754 & 0.832 & 0.724 \\	
		& Cross-Entropy+ \cite{good-closed-set} & 0.768 & 0.859 & 0.735 \\
		\cmidrule{2-5}
		& CFAN-OSFGR & \textbf{0.772} & \textbf{0.860} & \textbf{0.754}  \\
		\midrule
		\multirow{9}{*}{SwinB} & Backbone & 0.856 & 0.905 & 0.800 \\
		%		STAN-Transformer & 0.888 & 0.934  \\
		%		Trans-ADH \cite{OSR_transformer1} & 0.868 & 0.914 & 0.811   \\
		& OpenHybrid & 0.857 & 0.910 & 0.805  \\
		& ARPL & 0.860 & 0.908 & 0.807  \\
		& Cross-Entropy+ & 0.864 & 0.906 & 0.808  \\
		& Trans-AUG \cite{OSR_transformer2} & 0.860 & 0.910 & 0.809   \\
		& MoEP-AE-OSR \cite{MoEP-AE} & 0.869 & 0.922 & 0.819  \\
		\cmidrule{2-5}
		& GFNet \cite{frequency_enhancing1} & 0.869 & 0.917 & 0.830 \\
		& FcaNet \cite{frequency_enhancing3} & 0.872 & 0.912 & 0.826  \\
		\cmidrule{2-5}
		& CFAN-OSFGR & \textbf{0.895} & \textbf{0.939} & \textbf{0.859} \\
		\bottomrule
	\end{tabular}
	\label{table: Stanford-Cars}
	\vspace{-0.5cm}
\end{table}

\begin{table}[t]\footnotesize
	\renewcommand\arraystretch{0.5}
	\centering
	\setlength{\abovecaptionskip}{0cm}  %段前
	\caption{Computational resources on Aircraft.}
	\begin{tabular}{m{0.9cm}<{\centering}m{2.7cm}<{\raggedright}m{0.9cm}<{\centering}m{0.8cm}<{\centering}m{0.8cm}<{\centering}}
		\toprule
		Backbone & Method & FLOPs & Params & FPS  \\
		\midrule
		\multirow{7}{*}{CNN} & OpenHybrid \cite{OpenHybrid} & 32.5G & 33.6M & 24 \\
		& OpenGAN \cite{OpenGAN} & 29.3G & 29.1M & 35 \\	
		& ARPL \cite{ARPL} & 22.1G & 24.0M & 41 \\
		& GCPL \cite{GCPL} & 21.6G & 23.8M & 42   \\
		& GMVAE-OSR \cite{GMVAE-OSR} & 25.2G & 32.1M & 39   \\
		& CAMV \cite{IEEE-Access} & 20.7G & 23.6M & 47 \\	
		& Cross-Entropy+ \cite{good-closed-set} & 20.8G & 23.6M & 46 \\
		\cmidrule{2-5}
		& CFAN-OSFGR & 42.7G & 28.1M & 13  \\
		\midrule
		\multirow{9}{*}{SwinB} & Backbone & 60.6G & 86.8M & 29  \\
		& OpenHybrid & 89.9G & 98.5M & 13  \\
		& ARPL & 65.3G & 87.0M & 21  \\
		& Cross-Entropy+ & 60.7G & 86.8M & 28  \\
		& Trans-AUG \cite{OSR_transformer2} & 60.8G & 86.8M & 27   \\
		& MoEP-AE-OSR \cite{MoEP-AE} & 62.8G & 91.6M & 23  \\
		\cmidrule{2-5}
		& GFNet \cite{frequency_enhancing1} & 61.2G & 88.3M & 24  \\
		& FcaNet \cite{frequency_enhancing3} & 60.8G & 89.5M & 23  \\
		\cmidrule{2-5}
		& CFAN-OSFGR & 86.1G & 90.3M & 12 \\
		\bottomrule
	\end{tabular}
	\label{table: computation}
	\vspace{-0.5cm}
\end{table}

\begin{table}[t]\footnotesize
	\renewcommand\arraystretch{0.5}
	\centering
	\setlength{\abovecaptionskip}{0cm} 
	\caption{Comparison of the macro-F1 scores under the \textit{cross-dataset setting}. Under this setting, Aircraft is used as the known-class dataset, while CUB and Stanford-Cars are used as the unknown-class datasets respectively.}
	\begin{tabular}{m{1.3cm}<{\centering}m{2.7cm}<{\raggedright}m{1cm}<{\centering}m{1.8cm}<{\centering}}
		\toprule
		Backbone & Method & CUB & Stanford-Cars \\
		\midrule
		\multirow{8}{*}{CNN} & OpenHybrid \cite{OpenHybrid} & 0.473 & 0.840 \\
		& OpenGAN \cite{OpenGAN} & 0.426 & 0.819 \\	
		& ARPL \cite{ARPL} & 0.469 & 0.831 \\
		& GCPL \cite{GCPL} & 0.435 & 0.812  \\
		& GMVAE-OSR \cite{GMVAE-OSR} & 0.454 & 0.826   \\
		& CAMV \cite{IEEE-Access} & 0.461 & 0.833 \\	
		& Cross-Entropy+ \cite{good-closed-set} & 0.482 & 0.869 \\
		\cmidrule{2-4}
		& CFAN-OSFGR & \textbf{0.491} & \textbf{0.878}   \\
		\midrule
		\multirow{9}{*}{SwinB} & Backbone & 0.497 & 0.886  \\
		& OpenHybrid & 0.499 & 0.875  \\
		& ARPL & 0.521 & 0.888  \\
		& Cross-Entropy+ & 0.503 & 0.891  \\
		& Trans-AUG \cite{OSR_transformer2} & 0.508 & 0.882 \\
		& MoEP-AE-OSR \cite{MoEP-AE} & 0.529 & 0.876 \\
		\cmidrule{2-4}
		& GFNet \cite{frequency_enhancing1} & 0.518 & 0.890  \\
		& FcaNet \cite{frequency_enhancing3} & 0.522 & 0.888  \\ 
		\cmidrule{2-4}
		& CFAN-OSFGR & \textbf{0.546} & \textbf{0.893} \\
		\bottomrule
	\end{tabular}
	\label{table: cross-dataset}
	\vspace{-0.1cm}
\end{table}

Tables \ref{table: Aircraft}, \ref{table: CUB}, and \ref{table: Stanford-Cars} report the evaluation results under the \textit{standard-dataset setting} on Aircraft, CUB and Stanford-Cars, respectively. Table \ref{table: computation} reports the computational resources of these methods on Aircraft. Two points can be seen from these tables:

(1) Transformer-based models achieve better results than CNN-based models in most cases, except in a few cases (\emph{e.g.} on Aircraft) where CNN-based models perform slightly better than transformer-based models. The main reason is that the multiple self-attention operations in transformers boost the model discriminability. Besides, transformer-based models are generally slower than CNN-based models, because the calculations and parameters of fully-connected layers in transformers are generally larger than those of convolutional layers in CNNs. 

(2) CFAN-OSFGR outperforms all the comparative methods with the same CNN or SwinB backbone in most cases. We need to point out that CFAN-OSFGR is relatively slower than other methods with the same backbone, mainly due to the feature transformation in the frequency domain as well as the temporal feature aggregation. Furthermore, we conduct an experiment for evaluating the OSFGR performance of different models with similar calculations and parameters, whose results show that CFAN-OSFGR still outperforms other models with similar overheads.

\begin{table}[t]\footnotesize
	\renewcommand\arraystretch{0.5}
	\centering
	\setlength{\abovecaptionskip}{0cm}
	\caption{Comparison of AUROC and ACC results on the coarse-grained datasets.}
	\begin{tabular}{m{0.9cm}<{\centering}m{2.1cm}<{\raggedright}m{0.4cm}<{\centering}m{1.3cm}<{\centering}m{0.4cm}<{\centering}m{1cm}<{\centering}}
		\toprule
		\multirow{2}{*}{Backbone} & \multirow{2}{*}{Method} & \multicolumn{2}{c}{CIFAR+10/+50} & \multicolumn{2}{c}{TinyImageNet}  \\
		\cmidrule(r){3-6}
		& & ACC & AUROC & ACC & AUROC  \\
		\midrule
		\multirow{8}{*}{CNN} & OpenHybrid & 0.937 & 0.962/0.955 & 0.632 & 0.793 \\
		& OpenGAN & 0.940 & \textbf{0.981/0.983} & 0.684 & \textbf{0.907} \\	
		& ARPL & 0.940 & 0.965/0.943 & 0.638 & 0.762 \\
		& GCPL & 0.945 & 0.951/0.946 & 0.643 & 0.759   \\
		& GMVAE-OSR & 0.952 & 0.952/0.947 & 0.729 & 0.782   \\
		& CAMV & 0.942 & 0.942/0.933 & 0.796 & 0.805 \\	
		& Cross-Entropy+ & 0.958 & 0.954/0.939 & 0.862 & 0.826 \\
		\cmidrule{2-6}
		& CFAN-OSFGR & \textbf{0.961} & 0.967/0.959 & \textbf{0.865} & 0.881   \\
		\midrule
		\multirow{9}{*}{SwinB} & Backbone & 0.981 & 0.959/0.963 & 0.904 & 0.909  \\
		& OpenHybrid & 0.981 & 0.978/0.980 & 0.925 & 0.934  \\
		& ARPL & 0.983 & 0.980/0.985 & 0.929 & 0.927   \\
		& Cross-Entropy+ & 0.984 & 0.965/0.971 & 0.913 & 0.916   \\
		& Trans-AUG & 0.983 & 0.982/0.986 & 0.947 & 0.942  \\
		& MoEP-AE-OSR & 0.983 & 0.976/0.975 & \textbf{0.965} & 0.952  \\
		\cmidrule{2-6}
		& GFNet \cite{frequency_enhancing1} & 0.984 & 0.980/0.982 & 0.951 & 0.943   \\
		& FcaNet \cite{frequency_enhancing3} & 0.986 & 0.981/0.977 & 0.955 & 0.939  \\
		\cmidrule{2-6}
		& CFAN-OSFGR & \textbf{0.990} & \textbf{0.994/0.991} & 0.963 & \textbf{0.955}  \\
		\bottomrule
	\end{tabular}
	\label{table: coarse_grained_OSR}
	\vspace{-0.3cm}
\end{table}

\begin{table*}[t]\footnotesize
	\renewcommand\arraystretch{0.5}
	\centering
	\setlength{\abovecaptionskip}{0cm}
	\caption{Ablation results on Aircraft by the proposed CFAN-OSFGR method with different configurations of the three modules.}
	\begin{tabular}{m{1.6cm}<{\centering}m{1.6cm}<{\centering}m{1.6cm}<{\centering}m{2.4cm}<{\centering}m{3.2cm}<{\centering}m{3.2cm}<{\centering}}
		\toprule
		Backbone & FVF module & CTA module & ACC & \makecell[c]{AUROC\\ (Easy/Medium/Hard)} & \makecell[c]{OSCR\\ (Easy/Medium/Hard)} \\
		\midrule
		\ding{52} & \ding{56} &  \ding{56} & 0.896 & 0.848/0.827/0.723 & 0.799/0.778/0.686  \\
		\ding{52} & \ding{52} &  \ding{56} & 0.913 & 0.885/0.879/0.824 & 0.843/0.827/0.781  \\
		\ding{52} & \ding{52} & \ding{52} & 0.917 & 0.913/0.892/0.839 & 0.864/0.845/0.803  \\
		\bottomrule
	\end{tabular}
	\label{table: modules}
	\vspace{-0.4cm}
\end{table*}

\textbf{Evaluation Under the \textit{Cross-Dataset Setting}.} The above results have demonstrated the effectiveness of the proposed CFAN-OSFGR method in cases where known-class images and unknown-class images are from the same dataset.
Here, we also evaluate the model performance under the \textit{cross-dataset setting} where the unknown-class images are from outlier datasets. Specifically, the model is trained with Aircraft, and tested with images from CUB and Stanford-Cars as the unknown-class images respectively, whose evaluation results are reported in Table \ref{table: cross-dataset}. As seen from this table, CFAN-OSFGR outperforms all the other methods significantly in most cases, demonstrating its cross-dataset generalization ability.

\vspace{0.1cm}
\noindent \normalsize{\textbf{4.3.2 Evaluation on the Coarse-Grained Datasets}}
\vspace{0.1cm}

The above results have demonstrated the effectiveness of the proposed CFAN-OSFGR method in handling open-set fine-grained images. Here, we further conduct an experiment for evaluating the effectiveness of CFAN-OSFGR in dealing with open-set coarse-grained images. Table \ref{table: coarse_grained_OSR} reports the evaluation results on the two coarse-grained datasets (CIFAR+10/+50 and TinyImageNet). As seen from this table, CFAN-OSFGR still achieves the best results in most cases or achieves the 2nd place. These results further demonstrate
%the superiority of CFAN-OSFGR without regard to whether the images are fine-grained or coarse-grained.
that in an open-set scenario, CFAN-OSFGR can not only effectively recognize fine-grained images, but also recognize coarse-grained images accurately.

%\vspace{0.1cm}
%\noindent \normalsize{\textbf{4.3.3 Evaluation on HFI/LFI}}
%\vspace{0.1cm}
%
%We also evaluate the OSFGR performance of CFAN-OSFGR on HFI/LFI from Aircraft, whose results are reported in Table \ref{table: HRI_LRI}. As seen from this table, CFAN-OSFGR boosts the model performance on both HFI and LFI, indicating that CFAN-OSFGR has better captured both high- and low-frequency information than both ResNet50 and SwinB.

\begin{figure*}[t]
	\begin{center}
		\setlength{\abovecaptionskip}{0.cm}
		\includegraphics[height=11.3cm,width=18cm]{{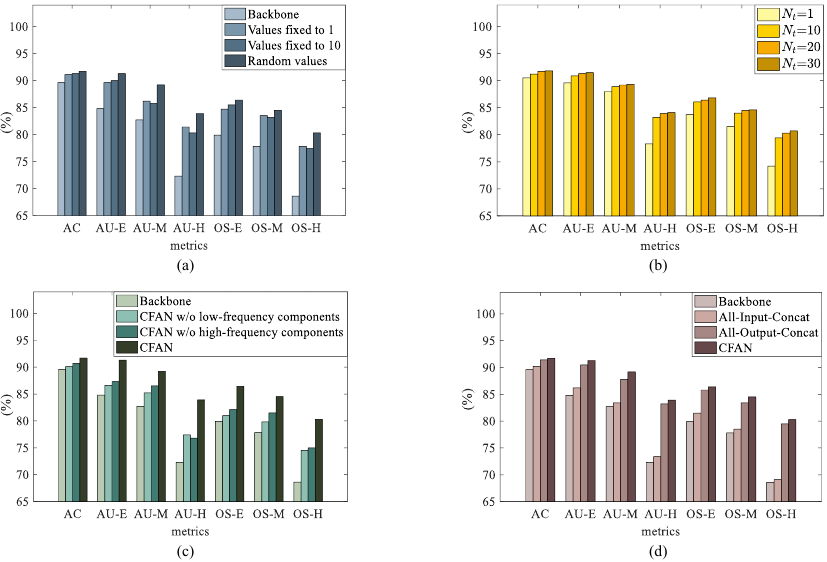}}
	\end{center}
	\vspace{-0.3cm}
	\caption{OSFGR results on Aircraft for analyzing the influence of: (a) randomizing the initial adjustable vectors $\boldsymbol{p}_h^1$ and $\boldsymbol{p}_l^1$ in the FVF module; (b) different numbers of template filters in the FVF module; (c) aggregating both high- and low-frequency components in the CTA module; (d) temporal aggregation in the CTA module. The three metrics, ACC, AUROC, and OSCR, are denoted as `AC', `AU', and `OS', respectively. The three difficulty modes, `Easy', `Medium', and `Hard', are abbreviated to `E', `M', and `H', respectively.}
	\label{fig: four_ablation}
	\vspace{-0.2cm}
\end{figure*}

\begin{table}[t]\footnotesize
	%	\vspace{-0.2cm}
	\renewcommand\arraystretch{0.5}
	\centering
	\setlength{\abovecaptionskip}{0cm}
	\caption{OSFGR results on Aircraft under the \textit{standard-dataset setting} by CFAN-OSFGR with random $\boldsymbol{p}_h^1$ and $\boldsymbol{p}_l^1$: the means and the standard deviations over 50 random trials are reported.}
	\begin{tabular}{m{0.8cm}<{\raggedright}m{1.0cm}<{\centering}m{2.6cm}<{\centering}m{2.6cm}<{\centering}}
		\toprule
		item & ACC & AUROC & OSCR \\
		\midrule
		%		report & 0.947 & 0.954/0.909/0.833 & 0.915/0.877/0.810  \\
		%		mean(exp) & 0.9467 & 0.9544/0.9093/0.8309 & 0.9152/0.8779/0.8083   \\
		mean & 0.9174 & 0.9134/0.8923/0.8386 & 0.8642/0.8449/0.8033   \\
		std & 0.0003 & 0.0003/0.0004/0.0008 & 0.0003/0.0004/0.0008  \\
		\bottomrule
	\end{tabular}
	\label{table: Aircraft_testing}
	\vspace{-0.3cm}
\end{table}

\subsection{Ablation Studies}  \label{section: ablation}

Here, we conduct extensive ablation studies for evaluating the effectiveness of CFAN-OSFGR more comprehensively. The following experiments are all implemented on Aircraft \cite{Aircraft} under the \textit{standard-dataset setting}, and are implemented on SwinB-based CFAN-OSFGR.
%The main results are reported and analyzed in the following subsections, and some additional results can be found in the supplementary material due to the limitation of space.

\vspace{0.1cm}
\noindent \normalsize{\textbf{4.4.1 Ablation Study on Modules}}
\vspace{0.1cm}

Firstly, we conduct an ablation study for analyzing the effect of the last two modules (\emph{i.e.}, the frequency-varying filtering (FVF) module and the complementary temporal aggregation (CTA) module) in the proposed CFAN. The corresponding results are reported in Table \ref{table: modules}. As seen from this table, the model performance improves with the two modules added to the model one by one, indicating that either of the two modules plays an important role in CFAN.

\vspace{0.1cm}
\noindent \normalsize{\textbf{4.4.2 Analysis of the FVF Module}}
\vspace{0.1cm}

%\begin{figure}[t]
%	\begin{center}
%		\setlength{\abovecaptionskip}{0.cm}
%		\includegraphics[height=3.8cm,width=6.8cm]{{randomness_training.pdf}}
%	\end{center}
%	\vspace{-0.5cm}
%	\caption{OSFGR results on Aircraft for analyzing the influence of randomizing the values in the adjustable vectors $\boldsymbol{p}_h^1$ and $\boldsymbol{p}_l^1$ in the FVF module at the training stage. The three metrics, ACC, AUROC, and OSCR, are denoted as `AC', `AU', and `OS', respectively. The three difficulty modes, `Easy', `Medium', and `Hard', are abbreviated to `E', `M', and `H', respectively.}
%	\label{fig: randomness}
%	\vspace{-0.3cm}
%\end{figure}

%\begin{figure}[t]
%	\begin{center}
%		\setlength{\abovecaptionskip}{0.cm}
%		\includegraphics[height=3.8cm,width=6.8cm]{{filter_numbers.pdf}}
%	\end{center}
%	\vspace{-0.5cm}
%	\caption{OSFGR results on Aircraft for analyzing the influence of the number of template filters $N_t$ in the FVF module. The three metrics, ACC, AUROC, and OSCR, are denoted as `AC', `AU', and `OS', respectively. The three difficulty modes, `Easy', `Medium', and `Hard', are abbreviated to `E', `M', and `H', respectively.}
%	\label{fig: filter_numbers}
%	\vspace{-0.3cm}
%\end{figure}

\textbf{Influence of Randomizing Initial Adjustable Vectors $\boldsymbol{p}_h^1$ and $\boldsymbol{p}_l^1$ on Performance.} As described in Sec. 3.1.2, the initial adjustable vectors $\boldsymbol{p}_h^1$ and $\boldsymbol{p}_l^1$ are initialized by randomly sampling from the uniform distribution $U(0,1)$. Here, we analyze the influence of such an initialization manner on model performance. 

Firstly, we always fix the values in $\boldsymbol{p}_h^1$ and $\boldsymbol{p}_l^1$ to $1$ and $10$ respectively, and the OSFGR results on Aircraft under the \textit{standard-dataset setting} are shown in Fig. \ref{fig: four_ablation} (a). Besides, we separately evaluate the proposed CFAN-OSFGR method with random $\boldsymbol{p}_h^1$ and $\boldsymbol{p}_l^1$ over 50 trials, and the corresponding results are shown in Table \ref{table: Aircraft_testing}. Two points can be seen from this subfigure and this table: (i) The average performance over 50 trials of randomizing $\boldsymbol{p}_h^1$ and $\boldsymbol{p}_l^1$ is better than that of fixing $\boldsymbol{p}_h^1$ and $\boldsymbol{p}_l^1$, demonstrating the effectiveness of the random initialization manner on $\boldsymbol{p}_h^1$ and $\boldsymbol{p}_l^1$; (ii) The standard deviations over 50 random trials are very small, demonstrating that the model is not sensitive to the random initialization manner on $\boldsymbol{p}_h^1$ and $\boldsymbol{p}_l^1$.

%\begin{figure}[t]
%	\begin{center}
%		\setlength{\abovecaptionskip}{0.cm}
%		\includegraphics[height=5.8cm,width=8.8cm]{{filter_numbers.pdf}}
%	\end{center}
%	\vspace{-0.3cm}
%	\caption{OSFGR results on Aircraft for analyzing the influence of the number of template filters $N_t$ in the FVF module. The three metrics, ACC, AUROC, and OSCR, are denoted as `AC', `AU', and `OS', respectively. The three difficulty modes, `Easy', `Medium', and `Hard', are abbreviated to `E', `M', and `H', respectively.}
%	\label{fig: filter_numbers}
%	\vspace{-0.2cm}
%\end{figure}

\textbf{Influence of Number of Template Filters.} Here we analyze the influence of the number of template filters $N_t$ in the FVF module. We evaluate CFAN-OSFGR with $N_t=\{ 1,10,20,30 \}$, and the results are shown in Fig. \ref{fig: four_ablation} (b). As seen from this subfigure, the performance varies slightly when $N_t$ varies in $[ 10,30 ]$, indicating that CFAN-OSFGR is not quite sensitive to $N_t$.

\vspace{0.1cm}
\noindent \normalsize{\textbf{4.4.3 Analysis of the CTA Module}}
\vspace{0.1cm}

%\begin{figure}[t]
%	%	\vspace{-0.3cm}
%	\begin{center}
%		\setlength{\abovecaptionskip}{0.cm}
%		\includegraphics[height=4.0cm,width=6.8cm]{{streams.pdf}}
%	\end{center}
%	\vspace{-0.5cm}
%	\caption{OSFGR results on Aircraft for analyzing the influence of aggregating both high- and low-frequency components in the CTA module. The three metrics, ACC, AUROC, and OSCR, are denoted as `AC', `AU', and `OS', respectively. The three difficulty modes, `Easy', `Medium', and `Hard', are abbreviated to `E', `M', and `H', respectively.}
%	\label{fig: aggregation}
%	\vspace{-0.3cm}
%\end{figure}

%\begin{figure}[t]
%	\begin{center}
%		\setlength{\abovecaptionskip}{0.cm}
%		\includegraphics[height=4.0cm,width=6.8cm]{{aggregation.pdf}}
%	\end{center}
%	\vspace{-0.5cm}
%	\caption{OSFGR results on Aircraft for analyzing the influence of temporal aggregation in the CTA module. The three metrics, ACC, AUROC, and OSCR, are denoted as `AC', `AU', and `OS', respectively. The three difficulty modes, `Easy', `Medium', and `Hard', are abbreviated to `E', `M', and `H', respectively.}
%	\label{fig: temporal}
%	\vspace{-0.4cm}
%\end{figure}

\textbf{Influence of Aggregating High- and Low-Frequency Components.} Here, we analyze the influence of aggregating the two time series of components (\emph{i.e.}, the time series of high-frequency components and the time series of low-frequency components) in the CTA module. Specifically, we train two additional models based on CFAN-OSFGR where the discriminative feature is obtained only from the high-frequency components by $LSTM_h$ or the low-frequency components by $LSTM_l$, whose results are shown in Fig. \ref{fig: four_ablation} (c). As shown in this subfigure, either the high- or the low-frequency component boosts the model performance to some extent but not significantly, and aggregating both time series provides significant performance improvements in most cases, especially under the AUROC and OSCR metrics. These results demonstrate the effectiveness of aggregating both high- and low-frequency information in boosting the generalization ability of the model.

\begin{table}[t]\footnotesize
	\renewcommand\arraystretch{0.5}
	\centering
	\setlength{\abovecaptionskip}{0cm}
	\setlength{\belowcaptionskip}{-0.2cm}
	\caption{OSFGR results on Aircraft by utilizing the features outputted from different stages of the SwinB backbone.}
	\begin{tabular}{m{0.8cm}<{\raggedright}m{1.0cm}<{\centering}m{2.6cm}<{\centering}m{2.6cm}<{\centering}}
		\toprule
		Stage & ACC & AUROC & OSCR  \\
		\midrule
		1 & 0.866 & 0.851/0.869/0.794 & 0.807/0.818/0.738  \\
		2 & 0.871 & 0.857/0.880/0.806 & 0.819/0.823/0.754  \\
		3 & 0.893 & 0.861/0.887/0.830 & 0.825/0.839/0.782  \\
		4 & 0.917 & 0.913/0.892/0.839 & 0.864/0.845/0.803  \\
		%backbone  & 0.896 & 0.848/0.827/0.723 & 0.799/0.778/0.686  \\
		\bottomrule
	\end{tabular}
	\label{table: backbone_layers}
	\vspace{-0.5cm}
\end{table}

\begin{table*}[t]\footnotesize
	\renewcommand\arraystretch{0.5}
	\centering
	\setlength{\abovecaptionskip}{0cm}
	\setlength{\belowcaptionskip}{0.5cm}
	\caption{OSFGR results and the computational resources on Aircraft by SwinB (Larger), Cross-Entropy+ (Larger), and the proposed CFAN-OSFGR method.}
	%	\vspace{1pt}
	\begin{tabular}{m{3.3cm}<{\raggedright}m{1.4cm}<{\centering}m{3.3cm}<{\centering}m{3.3cm}<{\centering}m{1.2cm}<{\centering}m{1.2cm}<{\centering}m{1.2cm}<{\centering}}
		\toprule
		Model & ACC & \makecell[c]{AUROC\\ (Easy/Medium/Hard)} & \makecell[c]{OSCR\\ (Easy/Medium/Hard)} & FLOPs & Params & FPS  \\
		\midrule
		SwinB (Larger) & 0.901 & 0.852/0.836/0.735 & 0.819/0.783/0.689 & 86.6G & 119.4M & 25  \\
		Cross-Entropy+ (Larger) & 0.906 & 0.864/0.858/0.743 & 0.825/0.804/0.690 & 86.8G & 121.2M & 23 \\
		CFAN-OSFGR & 0.917 & 0.913/0.892/0.839 & 0.864/0.845/0.803 & 86.1G & 90.3M & 12  \\
		\bottomrule
	\end{tabular}
	\label{table: larger}
	\vspace{-0.3cm}
\end{table*}

\textbf{Influence of Temporal Aggregation.} Moreover, we analyze the influence of temporal aggregation in the CTA module. Specifically, we compare the proposed temporal aggregation with other two aggregation strategies: (i) concatenating the $2 \cdot N_f$ input feature components in the two time series (denoted as `All-Input-Concat'), and (ii) concatenating the $2 \cdot N_f$ Hidden states at all moments (denoted as `All-Output-Concat'), whose results are shown in Fig. \ref{fig: four_ablation} (d). As shown in this subfigure, `All-Input-Concat' slightly improves the backbone model performance since it also aggregates high- and low-frequency information, but is inferior to either `All-Output-Concat' or CFAN. Besides, the results of `All-Output-Concat' are slightly lower than those of CFAN since aggregating the Hidden states at earlier moments weakens the effect of temporal modeling. All these results demonstrate the effectiveness of the temporal aggregation strategy, which also provides an enlightening insight into modeling the temporal dependence of an image feature for boosting the feature discriminability. 

%In addition, we also analyze the influence of $\boldsymbol{p}_h^1$ and $\boldsymbol{p}_l^1$ on evaluation, the influence of the number of high/low-pass filters $N_f$ in the FVF module, and the influence of the number of template filters $N_t$ in the FVF module, which can be found in the supplementary material.

\subsection{Influence of Different Layers of Backbone}  \label{section: backbone_layers}

%\begin{table}[t]\footnotesize
%	\renewcommand\arraystretch{0.5}
%	\centering
%	\setlength{\abovecaptionskip}{0cm}
%	\setlength{\belowcaptionskip}{-0.2cm}
%	\caption{OSFGR results on CUB under the AUROC and OSCR metrics in the `Hard' mode for analyzing the influence of different backbones.}
%	\begin{tabular}{m{0.8cm}<{\raggedright}m{1.0cm}<{\centering}m{2.6cm}<{\centering}m{2.6cm}<{\centering}}
%		\toprule
%		Stage & ACC & AUROC & OSCR  \\
%		\midrule
%		1 & 0.933 & 0.941/0.888/0.781 & 0.893/0.850/0.754  \\
%		2 & 0.933 & 0.946/0.894/0.789 & 0.897/0.855/0.762  \\
%		3 & 0.944 & 0.949/0.897/0.824 & 0.910/0.867/0.802  \\
%		4 & 0.950 & 0.959/0.899/0.828 & 0.921/0.870/0.806  \\
% backbone& 0.949 & 0.945/0.875/0.804 & 0.908/0.848/0.781  \\
%		\bottomrule
%	\end{tabular}
%	\label{table: backbone_layers}
%	\vspace{-0.4cm}
%\end{table}

Here, we evaluate the impacts of different layers (1st, 2nd, and 3rd stages) of the swin transformer backbone. Specifically, we take the features outputted from different stages of the SwinB backbone as the inputs, and input them to FVF and CTA modules. The OSFGR results on Aircraft are reported in Table \ref{table: backbone_layers}. As seen from this table, operating on the final stage of backbone achieves the best results, probably because that more self-attention layers could lead the model to pay more accurate attention on objects.

\subsection{Evaluation of Models With Similar Overheads}  \label{section: similar_overheads}

Here, we conduct an experiment for evaluating the OSFGR performance of different models with similar calculations and parameters. Specifically, we obtain a larger transformer backbone model (enlarged by adding W-MSA/SW-MSA blocks and upsampling operation to SwinB, called SwinB (Larger)) and a modified Cross-Entropy+ model (modified by replacing the original backbone with a larger transformer backbone, called Cross-Entropy+ (Larger)) which have similar parameters and calculations to SwinB-based CFAN-OSFGR. Then we evaluate their OSFGR performance on Aircraft, whose results are reported in Table \ref{table: larger}. As seen from this table, CFAN still significantly outperforms both SwinB (Larger) and Cross-Entropy+ (Larger), even with similar parameters and calculations, which further demonstrates the effectiveness of the proposed CFAN-OSFGR method.

\subsection{Visualization}  \label{section: visualization}

\begin{figure}[t]
	\begin{center}
		\setlength{\abovecaptionskip}{0.cm}
		\includegraphics[height=7cm,width=9cm]{{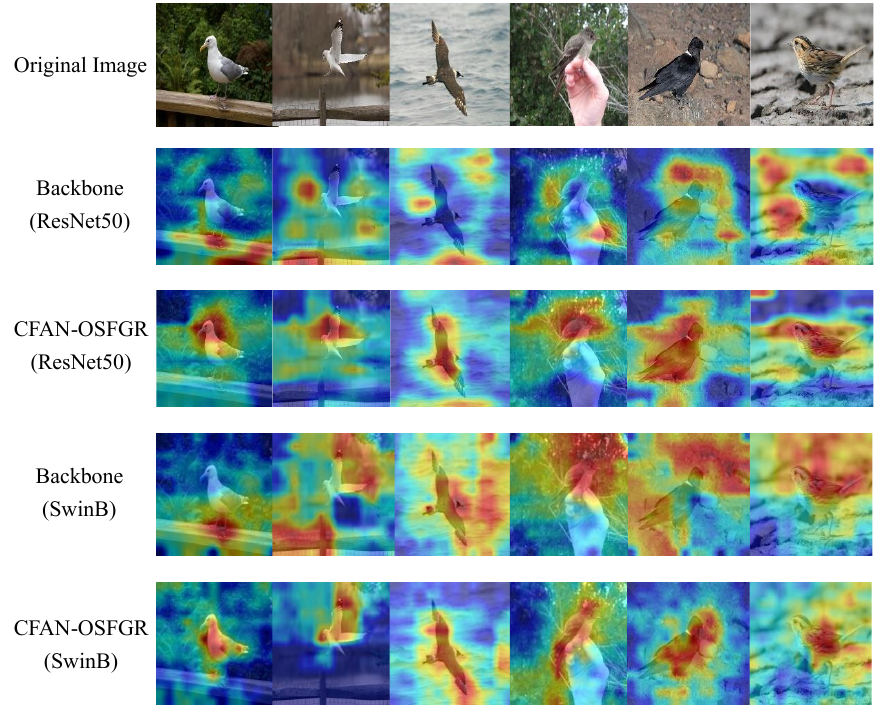}}
	\end{center}
	\vspace{-0.3cm}
	\caption{Heatmaps on six CUB images obtained by ResNet50 backbone (the second row), ResNet50-based CFAN-OSFGR (the third row), SwinB backbone (the fourth row), and SwinB-based CFAN-OSFGR (the fifth row). The attentions in red regions are strongest, while the attentions in yellow, green, and blue regions decrease by degrees.}
	\label{fig: visualization}
	\vspace{-0.2cm}
\end{figure}

Here, we visualize the heatmaps on six images from the CUB dataset obtained by backbone method and CFAN-OSFGR in Fig. \ref{fig: visualization}, where these images are erroneously predicted by backbone method but are correctly predicted by CFAN-OSFGR. As seen from this figure, CFAN could better concern high/low-frequency information of the semantic objects (\emph{e.g.}, the unique contours and details of Pomarine Jaeger in the 3rd column/the whole area of glaucous-winged Gull in the 1st column), hence, CFAN would predict correctly rather than predict as a similar class (long-tailed Jaeger also with sea background/Herring gull also with similar feet predicted by SwinB). It is mainly because CFAN is learned to be aware of the changing process from random high/low-frequency to full-band features, hence, CFAN can capture the high/low-frequency information better. Since the feature extractor is also trainable, the learned features by CFAN-OSFGR could better represent the high/low-frequency components in fine-grained images.

%\vspace{0.1cm}
%\noindent \normalsize{\textbf{4.4.4 Influence of Different Backbones}}
%\vspace{0.1cm}
%
%
%\begin{table}[t]\small
%	\renewcommand\arraystretch{0.6}
%	\centering
%	\setlength{\abovecaptionskip}{0cm}
%	\setlength{\belowcaptionskip}{-0.2cm}
%	\caption{OSFGR results on CUB under the AUROC and OSCR metrics in the `Hard' mode for analyzing the influence of different backbones.}
%	\begin{tabular}{m{4.2cm}<{\raggedright}m{1.2cm}<{\centering}m{1.2cm}<{\centering}}
%		\toprule
%		Model & AUROC & OSCR  \\
%		\midrule
%		VGG16 (Backbone) & 0.637 & 0.529  \\
%		VGG16 (CFAN-OSFGR) & 0.743 & 0.665  \\
%		\midrule
%		ResNet50 (Backbone) & 0.758 & 0.701  \\
%		ResNet50 (CFAN-OSFGR) & 0.795 & 0.732  \\
%		\midrule
%		SwinB (Backbone) & 0.804 & 0.781   \\
%		SwinB (CFAN-OSFGR) & 0.833 & 0.810  \\
%		\bottomrule
%	\end{tabular}
%	\label{table: different_backbones}
%	\vspace{-0.4cm}
%\end{table}
%
%
%Furthermore, we analyze the influence of different backbones in the proposed CFAN-OSFGR method. In addition to SwinB, we conduct experiments on VGG16 \cite{VGG} and ResNet50 \cite{ResNet} respectively, whose results are reported in Table \ref{table: different_backbones}. As seen from this table, the proposed CFAN-OSFGR method brings AUROC improvements of $10.6\%$, $3.7\%$, $2.9\%$ and OSCR improvements of $13.6\%$, $3.1\%$, $2.9\%$ for VGG16, ResNet50, SwinB, respectively. All these results demonstrate that the proposed CFAN is promising to be used as a stronger backbone for feature extraction, since it could achieve an effective balance between high- and low-frequency components.

\section{Conclusion}  \label{section: conclusion}

In this paper, we propose the complementary frequency-varying awareness network, CFAN, which learns discriminative features that could effectively capture and make use of both high- and low-frequency feature information from fine-grained images via three sequential modules: the feature extraction module, the frequency-varying filtering module where the frequency-adjustable filter is explored, and the complementary temporal aggregation module. Furthermore, the CFAN-OSFGR method is introduced for handling the open-set fine-grained image recognition task based on the proposed CFAN. Extensive experimental results demonstrate the effectiveness of CFAN-OSFGR. 

It is noted that the proposed method aims to better capture both high-frequency and low-frequency components in image features, there may still be imbalance between high-frequency information and low-frequency information in the learnt features. Hence, in our future work, we will further explore a feature aggregation strategy for balancing the high-frequency and low-frequency components in image features.

%A future research is to explore new techniques to exploit image information in temporal and frequency domains for learning more effective feature representations in other computer vision and pattern recognition tasks. 

{\small
\bibliographystyle{IEEEtran}
%\bibliography{egbib}
\bibliography{egbib_20221020}
}

\end{document}